\documentclass[letterpaper]{article} 
\usepackage{aaai23}  
\usepackage{times}  
\usepackage{helvet}  
\usepackage{courier}  
\usepackage[hyphens]{url}  
\usepackage{graphicx} 
\usepackage{natbib}  
\usepackage{caption} 
\usepackage{algorithm}
\usepackage{algorithmic}

\usepackage{newfloat}
\usepackage{listings}

\DeclareCaptionStyle{ruled}{labelfont=normalfont,labelsep=colon,strut=off} 
\floatstyle{ruled}
\newfloat{listing}{tb}{lst}{}
\floatname{listing}{Listing}
\usepackage[T5,T1]{fontenc}
\usepackage{combelow}
\usepackage[utf8]{inputenc}

\DeclareTextSymbolDefault{\ohorn}{T5}
\DeclareTextSymbolDefault{\uhorn}{T5}

\usepackage{amsmath}
\usepackage{amsfonts}       
\usepackage{amssymb}
\usepackage{amsthm}
\usepackage{enumerate}
\usepackage{enumitem,kantlipsum}

\usepackage{url}

\usepackage{cleveref}
\usepackage{mathtools}
\usepackage{dsfont}
\usepackage{xspace}
\usepackage[disable]{todonotes}
\usepackage{caption}
\usepackage{multirow}

\usepackage{comment}
\usepackage{booktabs}
\usepackage{tikz}
\usepackage{ltablex}
\usetikzlibrary{bayesnet,calc,matrix,positioning,shapes,backgrounds}

\definecolor{lightgray}{gray}{0.85}
\definecolor{lightlightgray}{gray}{0.9}
\definecolor{C1}{HTML}{1F77B4}
\definecolor{C2}{HTML}{FF7F0E}
\definecolor{C3}{HTML}{2CA02C}
\definecolor{C4}{HTML}{D62728}
\definecolor{C5}{HTML}{9467BD}

\crefname{section}{\S}{\S\S}
\Crefname{section}{\S}{\S\S}
\crefname{table}{Tab.}{}
\crefname{figure}{Fig.}{}
\crefname{algorithm}{Algorithm}{}
\crefname{equation}{eq.}{}
\crefname{appendix}{App.}{}
\crefname{thm}{Theorem}{}
\crefname{prop}{Proposition}{}
\crefname{cor}{Corollary}{}
\crefname{observation}{Observation}{}
\crefname{assumption}{Assumption}{}
\crefformat{section}{\S#2#1#3}

\usepackage{tikz}

\makeatletter
\newcommand*\iftodonotes{\if@todonotes@disabled\expandafter\@secondoftwo\else\expandafter\@firstoftwo\fi}  
\makeatother
\newcommand{\noindentaftertodo}{\iftodonotes{\noindent}{}}
\newcommand{\note}[4][]{\todo[author=#2,color=#3,size=\scriptsize,fancyline,caption={},#1]{#4}} 

\definecolor{dandelion}{HTML}{FFD464}
\newcommand{\lucas}[2][]{\note[#1]{lucas}{dandelion!60}{#2}}
\newcommand{\Lucas}[2][]{\lucas[inline,#1]{#2}\noindentaftertodo}

\newcommand{\karolina}[2][]{\note[#1]{karolina}{orange!10}{#2}}

\DeclareMathOperator*{\argmax}{argmax}

\newcommand{\R}{\mathbb{R}}

\newcommand{\SetSize}[1]{\lvert #1\rvert}

\def\calD{{\mathcal{D}}}

\def\vh{{\boldsymbol{h}}}

\def\vtheta{{\boldsymbol{\theta}\xspace}}

\def\vphi{{\boldsymbol{\phi}\xspace}}

\let\originalleft\left
\let\originalright\right
\renewcommand{\left}{\mathopen{}\mathclose\bgroup\originalleft}
\renewcommand{\right}{\aftergroup\egroup\originalright}

\newcommand{\sqr}[1]{\left[#1\right]}

\newcommand{\MI}{\mathrm{MI}}

\newcommand{\expectq}{\mathbb{E}_{\qphi}}

\newcommand{\bert}{BERT\xspace}

\newcommand{\vhC}{{\vh_C}}
\newcommand{\vhi}[1]{{\vh}^{(#1)}}
\newcommand{\vhn}{{\vh}^{(n)}}
\newcommand{\vhCn}{{\vh}^{(n)}_C}

\newcommand{\pii}[1]{{\pi}^{(#1)}}
\newcommand{\pin}{\pii{n}}

\newcommand{\NMI}{NMI\xspace}
\newcommand{\ent}{\mathrm{H}}

\newcommand{\ptheta}{p_{\vtheta}}

\newcommand{\qphi}{q_{\vphi}}
\newcommand{\qphik}{q^{\mathrm{size}}_{\vphi}}
\newcommand{\qphiC}{q^{\mathrm{CP}}_{\vphi}}

\newcommand{\mask}{\textsc{mask}\xspace}

\newcommand{\grad}{\nabla}
\newcommand{\gradTheta}{\grad_{\vtheta}}
\newcommand{\gradPhi}{\grad_{\vphi}}

\newcommand{\prop}[1]{\textsc{#1}}

\newcommand{\ZCP}{Z^\text{CP}}
\newcommand{\XX}{29\xspace}

\newcommand{\lowerbound}{\textsc{Linear}\xspace}
\newcommand{\upperbound}{\textsc{Upper Bound}\xspace}
\newcommand{\condpoisson}{\textsc{Conditional Poisson}\xspace}
\newcommand{\poisson}{\textsc{Poisson}\xspace}
\newcommand{\qda}{\textsc{Gaussian}\xspace}

\newcommand{\breakslash}{/\allowbreak\xspace}

\definecolor{greencolorname}{rgb}{0.4, 0.76, 0.65}
\definecolor{lightgreencolorname}{rgb}{0.65, 0.85, 0.33}
\definecolor{orangecolorname}{rgb}{0.99, 0.55, 0.3}
\definecolor{purplecolorname}{rgb}{0.55, 0.63, 0.80}
\definecolor{pinkcolorname}{rgb}{1.0, 0.33, 0.64}

\definecolor{deepcolorA}{rgb}{0.4, 0.76, 0.65}
\definecolor{deepcolorB}{rgb}{0.99, 0.55, 0.3}
\definecolor{deepcolorC}{rgb}{0.55, 0.63, 0.80}

\definecolor{mlp1}{rgb}{0.7, 0.7, 0.7}
\definecolor{mlp2}{rgb}{1, 0.85, 0.18}

\newcommand{\lowerboundC}{\textcolor{pinkcolorname}{\lowerbound}\xspace}

\newcommand{\poissonC}{\textcolor{orangecolorname}{\poisson}\xspace}
\newcommand{\condpoissonC}{\textcolor{purplecolorname}{\condpoisson}\xspace}
\newcommand{\qdaC}{\textcolor{greencolorname}{\qda}\xspace}

\title{A Latent-Variable Model for Intrinsic Probing}
\author{
    Karolina Sta\'nczak\equalcontrib,\textsuperscript{\rm 1}
    Lucas Torroba Hennigen\equalcontrib,\textsuperscript{\rm 2}
    Adina Williams,\textsuperscript{\rm 3} \\
    Ryan Cotterell,\textsuperscript{\rm 4}
    Isabelle Augenstein\textsuperscript{\rm 1}
}
\affiliations{
    \textsuperscript{\rm 1}University of Copenhagen 
    \textsuperscript{\rm 2}Massachusetts Institute of Technology
    \textsuperscript{\rm 3}Meta AI 
    \textsuperscript{\rm 4}ETH Z\"urich\\
    
    {\tt {ks@di.ku.dk}} \quad {\tt {lucastor@mit.edu}} \quad {\tt {adinawilliams@meta.com}} \\ 
{\tt {rcotterell@inf.ethz.ch}} \quad {\tt augenstein@di.ku.dk} 
}
\title{A Latent-Variable Model for Intrinsic Probing}

\usepackage{bibentry}

\begin{document}

\maketitle

\begin{abstract}
The success of pre-trained contextualized representations has prompted researchers to analyze them for the presence of linguistic information. 
Indeed, it is natural to assume that these pre-trained representations do encode some level of linguistic knowledge as they have brought about large empirical improvements on a wide variety of NLP tasks, which suggests they are learning true linguistic generalization.
In this work, we focus on intrinsic probing, an analysis technique where the goal is not only to identify whether a representation encodes a linguistic attribute but also to pinpoint \textit{where} this attribute is encoded.
We propose a novel latent-variable formulation for constructing intrinsic probes and derive a tractable variational approximation to the log-likelihood.
Our results show that our model is versatile and yields tighter mutual information estimates than two intrinsic probes previously proposed in the literature.
Finally, we find empirical evidence that pre-trained representations 
develop a cross-lingually entangled notion of morphosyntax.\footnote{Code is available at: \url{https://github.com/copenlu/flexible-probing}.}
\end{abstract}

\section{Introduction}

There have been considerable improvements to the quality of pre-trained contextualized representations in recent years~\citep[e.g.,][]{petersDeepContextualizedWord2018,devlinBERTPretrainingDeep2019,t5}.
These advances have sparked an interest in understanding what linguistic information may be lurking within the representations themselves~\citep[\emph{inter alia}]{poliakCollectingDiverseNatural2018,zhang-bowman-2018-language,rogers-etal-2020-primer}.
One philosophy that has been proposed to extract this information is called probing, the task of training an external classifier to predict the linguistic property of interest directly from the representations.
The hope of probing is that it sheds light onto how much linguistic knowledge is present in representations and, perhaps, how that information is structured.
Probing has grown to be a fruitful area of research, with researchers probing for morphological~\citep{tang-etal-2020-understanding-pure, acs-etal-2021-subword}, syntactic~\citep{voita-titov-2020-information, hall-maudslay-etal-2020-tale, acs-etal-2021-subword}, and semantic~\citep{vulic-etal-2020-probing,tang-etal-2020-understanding-pure} information.\looseness=-1
In this paper, we focus on one type of probing known as intrinsic probing \citep{dalviWhatOneGrain2019,intrinsic}, a subset of which specifically aims to ascertain how  information is structured within a representation.
This means that we are not solely interested in determining whether a network encodes the tense of a verb, but also in pinpointing exactly \emph{which} neurons in the network are responsible for encoding the property.
Unfortunately, the na{\"i}ve formulation of intrinsic probing requires one to test all possible combinations of neurons, which is intractable even for the smallest representations used in modern-day NLP.
For example, analyzing all combinations of 768-dimensional \bert representations would require training $2^{768}$ probes, one for each combination of neurons, which far exceeds the estimated number of atoms in the observable universe.


To obviate this difficulty, we introduce a novel latent-variable probe for intrinsic probing.
Our core idea, instead of training a different probe for each subset of neurons, is to introduce a subset-valued latent variable. 
We approximately marginalize over the latent subsets using variational inference.
Training the probe in this manner results in a set of parameters that work well across all possible subsets. 
We propose two variational families to model the posterior over the latent subset-valued random variables, both based on common sampling designs: Poisson sampling, which selects each neuron based on independent Bernoulli trials, and conditional Poisson sampling, which first samples a fixed number of neurons from a uniform distribution and then a subset of neurons of that size \citep{lohr2019sampling}.
Conditional Poisson sampling offers the modeler more control over the distribution over subset sizes; they may pick the parametric distribution themselves.


We compare both variants to the two main intrinsic probing approaches we are aware of in the literature (\S\ref{sec:results}).
To do so, we train probes for \XX morphosyntactic properties across 6 languages\footnote{Arabic, English, Finnish, Polish, Portuguese, and Russian} from the Universal Dependencies (UD; \citealt{ud-2.1}) treebanks.
We show that, in general, both variants of our method yield tighter estimates of the mutual information, though the model based on conditional Poisson sampling yields slightly better performance.
This suggests that they are better at quantifying the informational content encoded in m-\bert representations~\citep{devlinBERTPretrainingDeep2019}.
We make two typological findings when applying our probe. We show that there is a difference in how information is structured depending on the language with certain language--attribute pairs requiring more dimensions to encode relevant information.
We also analyze whether neural representations are able to learn cross-lingual abstractions from multilingual corpora. We confirm this statement and observe a strong overlap in the most informative dimensions, especially for number and gender. 
In an additional experiment, we show that our method supports training deeper probes (\cref{sec:results-deeper}), though the advantages of non-linear probes over their linear counterparts are modest.

\section{Intrinsic Probing}
\label{sec:background}


The success behind pre-trained contextual representations such as \bert~\citep{devlinBERTPretrainingDeep2019} suggests that they may offer a continuous analogue of the discrete structures in language, such as morphosyntactic attributes number, case, or tense. 
Intrinsic probing aims to recognize the parts of a network (assuming they exist) which encode such structures.
In this paper, we operate exclusively at the level of the neuron---in the case of BERT, this is one component of the 768-dimensional vector the model outputs.
However, our approach can easily generalize to other settings, e.g., the layers in a transformer or filters of a convolutional neural network.
Identifying individual neurons responsible for encoding linguistic features of interest has previously been shown to increase model transparency~\citep{bauIdentifyingControllingImportant2019}.
In fact, knowledge about which neurons encode certain properties has also been employed to mitigate potential biases~\citep{vigInvestigatingGB2020}, 
for controllable text generation~\citep{bauIdentifyingControllingImportant2019},
and to analyze the linguistic capabilities of language models~\citep{lakretzEmergenceNumberSyntax2019}.\looseness=-1

To formally describe our intrinsic probing framework, we first introduce some notation.
We define $\Pi$ to be the set of values that some property of interest can take, e.g., $\Pi = \{\prop{Singular}, \prop{Plural}\}$ for the morphosyntactic number attribute.
Let $\calD = \{ (\pi^{(n)}, \vh^{(n)}) \}_{n=1}^N$ be a dataset of label--representation pairs: $\pi^{(n)} \in \Pi$ is a linguistic property and $\vh^{(n)} \in \R^d$ is a representation.
Additionally, let $D$ be the set of all neurons in a representation; in our setup, it is an integer range.
In the case of \bert, we have $D = \{1, \ldots, 768\}$.
Given a subset of dimensions $C \subseteq D$, we write $\vh_C$ for the subvector of $\vh$ which contains only the dimensions present in $C$.

Let $\ptheta(\pin \mid \vhCn)$ be a probe---a classifier trained to predict $\pin$ from a subvector $\vhCn$. 
In intrinsic probing, our goal is to find the size $k$ subset of neurons $C \subseteq D$ which are most informative about the property of interest.
This may be written as the following combinatorial optimization problem~\citep{intrinsic}:
\begin{equation}\label{eq:optimization}
    C^\star = \argmax_{\substack{C \subseteq D, \\ |C| = k}} \sum_{n=1}^N \log \ptheta\left(\pi^{(n)} \mid \vh^{(n)}_C\right)
\end{equation} 
To exhaustively solve \Cref{eq:optimization}, we would have to train a probe $\ptheta\left(\pi \mid \vh_C\right)$ for every one of the exponentially many subsets $C \subseteq D$ of size $k$. 
Thus, exactly solving \cref{eq:optimization} is infeasible, and we are forced to rely on an approximate solution, e.g., greedily selecting the dimension that maximizes the objective.
However, greedy selection alone is not enough to make solving \cref{eq:optimization} manageable; because we must \emph{retrain} $\ptheta\left(\pi \mid \vh_C\right)$
for \emph{every} subset $C \subseteq D$ considered during the greedy selection procedure, i.e., we would end up training $\mathcal{O}\left(k\,|D|\right)$ classifiers.
As an example, consider what would happen if one used a greedy selection scheme to find the 50 most informative dimensions for a property on 768-dimensional \bert representations. To select the first dimension, one would need to train 768 probes. 
To select the second dimension, one would train an additional 767, and so forth. After 50 dimensions, one would have trained 37893 probes.
To address this problem, our paper introduces a latent-variable probe, which identifies a $\vtheta$ that can be used for any combination of neurons under consideration allowing a greedy selection procedure to work in practice.

\section{A Latent-Variable Probe}
\label{sec:method}

The technical contribution of this work is a novel latent-variable model for intrinsic probing.
Our method starts with a generic probabilistic probe
$\ptheta(\pi \mid C, \vh)$
which predicts a linguistic attribute $\pi$ given
a subset $C$ of the hidden dimensions;
$C$ is then used to subset $\vh$ into $\vhC$.
To avoid training a unique probe $\ptheta(\pi \mid C, \vh)$ for every possible subset $C\subseteq D$, we propose to integrate a prior over subsets $p(C)$ into the model and then to marginalize out all possible subsets of neurons:
\begin{align}\label{eq:joint}
    \ptheta(\pi \mid \vh) &= \sum_{C \subseteq D} \ptheta(\pi \mid C, \vh)\,p(C) 
\end{align}
Due to this marginalization, our likelihood is \emph{not} dependent on any specific subset of neurons $C$.
Throughout this paper, we opted for a non-informative, uniform prior $p(C)$, but other distributions are also possible.

Our goal is to estimate the parameters 
$\vtheta$.
We achieve this by maximizing
the log-likelihood of the training data
$\sum_{n=1}^N \log \sum_{C \subseteq D} \ptheta(\pi^{(n)}, C\mid \vhn)$
with respect to the parameters $\vtheta$.
Unfortunately, directly computing this involves a sum over all
possible subsets of $D$---a sum with an exponential number of summands. 
Thus, we resort to a variational approximation.
Let $\qphi(C)$ be a distribution over subsets, parameterized by parameters $\vphi$;
we will use $\qphi(C)$ to approximate the true posterior distribution. 
Then, the log-likelihood is lower-bounded as follows:
\begin{align}
    &\sum_{n=1}^N \log \sum_{C \subseteq D} \ptheta(\pi^{(n)}, C\mid \vhn) \label{eq:vb} \\
    &\ge  \sum_{n=1}^N\left( \expectq \sqr{\log \ptheta(\pii{n}, C \mid \vhi{n})} \hspace{-0.1cm} + \hspace{-0.1cm} \mathrm{H}(q)\right) \nonumber
\end{align}
which follows from Jensen's inequality, where $\mathrm{H}(\qphi)$ is the entropy of $\qphi$. 
The derivation of the variational lower bound is shown below:
\begin{align}
    &\sum_{n=1}^N \log \sum_{C \subseteq D} \ptheta(\pi^{(n)}, C\mid \vhn) \\
    \,\,&= \sum_{n=1}^N \log \sum_{C \subseteq D} \qphi(C) \frac{\ptheta(\pii{n}, C \mid \vhi{n})}{\qphi(C)} \nonumber\\
    \,\,&= \sum_{n=1}^N \log \expectq \sqr{\frac{\ptheta(\pii{n}, C \mid \vhi{n})}{\qphi(C)}} \nonumber\\
    \,\,&\ge \sum_{n=1}^N \expectq \sqr{\log \frac{\ptheta(\pii{n}, C \mid \vhi{n})}{\qphi(C)}} \label{eq:app-vb}  \\
    \,\,&=  \sum_{n=1}^N\left( \expectq \sqr{\log \ptheta(\pii{n}, C \mid \vhi{n})} + \mathrm{H}(q)\right) \nonumber
\end{align}

Our likelihood is general and can take the form of any objective function.
This means that we can use this approach to train intrinsic probes with any type of architecture amenable to gradient-based optimization, e.g., neural networks. However, in this paper, we use a linear classifier unless stated otherwise. Further, note that \cref{eq:vb} is valid for any choice of $\qphi$.
We explore two variational families for $\qphi$, each based on a common sampling technique. The first 
(herein \poisson) applies Poisson sampling \citep{hajekAsymptoticTheoryRejective1964}, which assumes each neuron to be subjected to an independent Bernoulli trial. 
The second one \citep[\condpoisson;][]{aires1999algorithms} corresponds to conditional Poisson sampling, which can be defined as conditioning a Poisson sample by a fixed sample size. 

\subsection{Parameter Estimation}\label{sec:learning}
As mentioned above, the exact computation of the log-likelihood is intractable due to the sum over all possible subsets of $D$. 
Thus, we optimize the variational bound presented in \cref{eq:vb}.
We optimize the bound through stochastic gradient descent with respect to the model parameters $\vtheta$ and the variational parameters $\vphi$, a technique known as stochastic variational inference~\citep{svi}.
However, one final trick is necessary, since the variational bound still includes a sum over all subsets in the first term:
\begin{align}
    \gradTheta \expectq &\sqr{\log \ptheta(\pii{n}, C \mid \vhi{n})} \\
    &\,\,= \expectq \sqr{ \gradTheta \log \ptheta(\pii{n}, C \mid \vhi{n}) } \nonumber \\
    &\,\,\approx \sum_{m=1}^M \sqr{ \gradTheta \log \ptheta(\pii{n}, C^{(m)} \mid \vhi{n}) }  \nonumber
\end{align}
where we take $M$ Monte Carlo samples to approximate the sum.
In the case of the gradient with respect to $\vphi$, we also have to apply the REINFORCE trick ~\citep{Williams1992SimpleSG}:
\begin{align}
    &\gradPhi \expectq \sqr{\log \ptheta(\pii{n}, C \mid \vhi{n})} \\
    &\,\,= \expectq \sqr{\log \ptheta(\pii{n}, C \mid \vhi{n}) \gradPhi \log \qphi(C)} \nonumber \\
    &\,\,\approx \sum_{m=1}^M \sqr{\log \ptheta(\pii{n}, C^{(m)} \mid \vhi{n}) \gradPhi \log \qphi(C)} \nonumber
\end{align}
where we again take $M$ Monte Carlo samples.
This procedure leads to an unbiased estimate of the gradient of the variational approximation.

\subsection{Choice of Variational Family $\qphi(C)$.}
We consider two choices of variational family $\qphi(C)$, both based on sampling designs \cite{lohr2019sampling}. 
Each defines a parameterized distribution over all subsets of $D$. 
\paragraph{Poisson Sampling.}
Poisson sampling is one of the simplest sampling
designs. 
In our setting, each neuron $d$ is given
a unique non-negative weight $w_d = \exp(\phi_d)$. 
This gives us the following parameterized distribution over subsets:
\begin{equation}\label{eq:coins}
    \qphi(C) = \prod_{d \in C} \frac{w_d}{1+w_d} \prod_{d \not\in C} \frac{1}{1 + w_d}
\end{equation}
The formulation in \cref{eq:coins} shows that taking a sample corresponds to $|D|$ independent coin flips---one for each neuron---where the probability of heads is $\frac{w_d}{1+w_d}$.
The entropy of a Poisson sampling may be computed in $\mathcal{O}\left(|D|\right)$ time:
\begin{equation}\label{eq:poisson-entropy}
     \ent(\qphi) = \log Z - \sum_{d = 1}^{\SetSize{D}} \frac{w_d}{1 + w_d} \log w_d
\end{equation}
where $\log Z = \sum_{d=1}^{\SetSize{D}} \log (1 + w_d)$.
The gradient of \cref{eq:poisson-entropy} may be computed automatically through backpropagation. 
Poisson sampling automatically modules the size of the sampled set $C \sim \qphi(\cdot)$ and we have the expected size $\mathds{E}\left[|C|\right] = \sum_{d=1}^{|D|} \frac{w_d}{1+w_d}$. 

\paragraph{Conditional Poisson Sampling.}
We also consider a variational family that factors
as follows:
\begin{equation}
    \qphi(C) = \underbrace{\qphiC(C \mid |C| = k)}_{\text{Conditional Poisson}}\,\qphik(k)
\end{equation}
In this paper, we take $\qphik(k) = \mathrm{Uniform}\left(D\right)$, but a more complex distribution, e.g., a Categorical, could be learned. 
We define $\qphiC(C \mid |C| = k)$ as a conditional Poisson sampling design.
Similarly to Poisson sampling, conditional Poisson sampling starts with a unique positive weight associated with every neuron $w_d = \exp(\phi_d)$.
However, an additional cardinality constraint is introduced. 
This leads to the following distribution:
\begin{equation}
    \qphiC(C) =\mathds{1}\left\{|C| = k\right\} \frac{\prod_{d \in C} w_d}{\ZCP}
\end{equation}
A more elaborate dynamic program which runs in $\mathcal{O}\left(k\,|D| \right)$ may be used to compute $\ZCP$ efficiently~\citep{aires1999algorithms}.
We may further compute the entropy $\mathrm{H}(\qphi)$ and its the gradient in $\mathcal{O}\left(|D|^2 \right)$
time using the expectation semiring~\cite{eisner-2002-parameter, li-eisner-2009-first}.
Sampling from $\qphiC$ can be done efficiently using quantities computed when running the dynamic program used to compute $\ZCP$~\citep{Kulesza_2012}.\footnote{
We use the semiring implementation by \citet{rushTorchStructDeepStructured2020}.}

\section{Experimental Setup}
\label{sec:experiment-setup}

Our setup is virtually identical to the morphosyntactic probing setup of \citet{intrinsic}.
This consists of first automatically mapping treebanks from UD v2.1~\citep{ud-2.1} to the UniMorph~\citep{mccarthyMarryingUniversalDependencies2018} schema.\footnote{We adopt the code available at: \url{https://github.com/unimorph/ud-compatibility}.}
Then, we compute multilingual \bert (m-\bert) representations\footnote{We use the implementation by \citet{wolfHuggingFaceTransformersStateoftheart2020}.} for every sentence in the UD treebanks.
After computing the m-\bert representations for the entire sentence, we extract representations for individual words in the sentence and pair them with the UniMorph morphosyntactic annotations.
We estimate our probes' parameters using the UD training set and conduct greedy selection to approximate the objective in \cref{eq:optimization} on the validation set; finally, we report the results on the test set, i.e., we test whether the set of neurons we found on the development set generalizes to held-out data.
Additionally, we discard values that occur fewer than 20 times across splits. 
When feeding $\vhC$ as input to our probes, we set any dimensions that are not present in $C$ to zero. We select $M=5$ as the number of Monte Carlo samples since we found this to work adequately in small-scale experiments. 
We compare the performance of the probes on \XX language--attribute pairs (listed in \cref{sec:app-attributes}).

Since the performance of a probe on a specific subset of dimensions is related to both the subset itself (e.g., whether it is informative or not) and the number of dimensions being evaluated (e.g., if a probe is trained to expect 768 dimensions as input, it might work best when few or no dimensions are filled with zeros), we sample 100 subsets of dimensions with 5 different possible sizes (we considered 10, 50, 100, 250, 500 dim.) and compare every model's performance on each of those subset sizes. 

\subsection{Baselines}\label{sec:baselines}
We compare our latent-variable probe against two other recently proposed intrinsic probing methods as baselines.
\begin{itemize}
\item \textbf{\citet{intrinsic}:} Our first baseline is a generative probe that models the joint distribution of representations and their properties $p(\vh, \pi) = p(\vh \mid \pi) \, p(\pi)$,
where the representation distribution $p(\vh \mid \pi)$ is assumed to be Gaussian.
\citet{intrinsic} report that a major limitation of this probe is that if certain dimensions of the representations are not distributed according to a Gaussian distribution, then probe performance will suffer.
\item \textbf{\citet{dalviWhatOneGrain2019}:} 
Our second baseline is a linear classifier, where dimensions not under consideration are zeroed out during evaluation \citep{dalviWhatOneGrain2019, durrani-etal-2020-analyzing}.\footnote{We note that they do not conduct intrinsic probing via dimension selection: Instead, they use the absolute magnitude of the weights as a proxy for dimension importance. In this paper, we adopt the approach of \citep{intrinsic} and use the performance-based objective in \cref{eq:optimization}.} 
Their approach is a special case of our proposed latent-variable model, where $\qphi$ is fixed so that on every training iteration the entire set of dimensions is sampled.
\end{itemize}

Additionally, we compare our methods to a na\"{i}ve approach, a probe that is re-trained for every set of dimensions under consideration selecting the dimension that maximizes the objective (herein \upperbound).\footnote{The \upperbound yields the tightest estimate on the mutual information, however as mentioned in \cref{sec:background}, this is unfeasible since it requires retraining for every different combination of neurons. For comparison, in English number, on an Nvidia RTX 2070 GPU, our \poisson, \qda, and \lowerbound experiments take a few minutes or even seconds to run, compared to \upperbound which takes multiple hours.}
Due to computational cost, we limit our comparisons with \upperbound to 6 randomly chosen morphosyntactic attributes,\footnote{English--Number, Portuguese--Gender and Noun Class, Polish--Tense, Russian--Voice, Arabic--Case, Finnish--Tense} each in a different language.

\subsection{Metrics}
\label{sec:metrics}

We compare our proposed method to the baselines above under two metrics: accuracy and mutual information (MI).
We report mutual information, which has recently been proposed as an evaluation metric for probes \citep{pimentelInformationtheoreticProbingLinguistic2020}. 
Here, mutual information (MI) is a function between a $\Pi$-valued random variable $P$ and a $\mathds{R}^{|C|}$-valued random variable $H_C$ over masked representations:
\begin{align}
    \MI(P; H_C) = \ent(P) - \ent(P \mid H_C)
\end{align}
where $\ent(P)$ is the inherent entropy of the property being probed and is constant with respect to $H_C$; $\ent(P \mid H_C)$ is the entropy over the property given the representations $H_C$. 
Exact computation of the mutual information is intractable; however, we can lower-bound the MI by approximating $\ent(P \mid H_C)$ using our probe's average negative log-likelihood: $-\frac{1}{N}\sum_{n=1}^N \log \ptheta(\pin \mid C, \vhn)$ on held-out data.
See \citet{brownEstimateUpperBound1992} for a derivation. 
We normalize the mutual information (\NMI) by dividing the MI by the entropy which turns it into a percentage and is, arguably, more interpretable. 
We refer the reader to \citet{gatesElementcentricClusteringComparison2019} for a discussion of the normalization of MI.


We also report accuracy which is a standard measure for evaluating probes as it is for evaluating classifiers in general. However, accuracy can be a misleading measure, especially on imbalanced datasets since it considers solely correct predictions.\looseness=-1

\subsection{What Makes a Good Probe?}\label{sec:how-to-compare}
Since we report a lower bound on the mutual information (\cref{sec:experiment-setup}), we deem the best probe to be the one that yields the tightest mutual information estimate, or, in other words, the one that achieves the highest mutual information estimate; this is equivalent to having the best cross-entropy on held-out data, which is the standard evaluation metric for language modeling.  

However, in the context of intrinsic probing, the topic of primary interest is what the probe reveals about the structure of the representations.
For instance, does the probe reveal that the information encoded in the embeddings is focalized or dispersed across neurons? 
Several prior works \citep[e.g.,][]{lakretzEmergenceNumberSyntax2019} focus on the single neuron setting, which is a special, very focal case.
To engage with this work, we compare probes not only with respect to their performance (MI and accuracy), but also with respect to the size of the subset of dimensions being evaluated, i.e., the size of set $C$.\looseness=-1

We acknowledge that there is a disparity between the quantitative evaluation we employ, in which probes are compared based on their MI estimates, and the qualitative nature of intrinsic probing, which aims to identify the substructures of a model that encode a property of interest.
However, it is non-trivial to evaluate fundamentally qualitative procedures in a large-scale, systematic, and unbiased manner.
Therefore, we rely on the quantitative evaluation metrics presented in \cref{sec:metrics}, while also qualitatively inspecting the implications of our probes.




\subsection{Training and Hyperparameter Tuning}
\label{sec:training}

We train our probes for a maximum of $2000$ epochs using the Adam optimizer~\citep{kingmaAdamMethodStochastic2015}. We add early stopping with a patience of $50$ as a regularization technique.
Early stopping is conducted by holding out 10\% of the training data; our development set is reserved for the greedy selection of subsets of neurons.
Our implementation is built with PyTorch~\citep{paszkePyTorchImperativeStyle2019}.
To execute a fair comparison with \citet{dalviWhatOneGrain2019}, we train all probes other than the Gaussian probe using ElasticNet regularization~\citep{zouRegularizationVariableSelection2005},
which consists of combining both $L_1$ and $L_2$ regularization, where the regularizers are weighted by tunable regularization coefficients $\lambda_1$ and $\lambda_2$, respectively.
We follow the experimental set-up proposed by \citet{dalviWhatOneGrain2019}, where we set $\lambda_1, \lambda_2 = 10^{-5}$ for all probes.
In a preliminary experiment, we performed a grid search over these hyperparameters to confirm that the probe is not very sensitive to the tuning of these values (unless they are extreme) which aligns with the claim presented in \citet{dalviWhatOneGrain2019}. For \qda, we take the MAP estimate, with a weak data-dependent prior~\citep[Chapter 4]{murphyMachineLearningProbabilistic2012}.
In addition, we found that a slight improvement in the performance of \poisson and \condpoisson was obtained by scaling the entropy term in \cref{eq:vb} by a factor of $0.01$.

\section{Results}
\label{sec:results}

In this section, we present the results of our empirical investigation. First, we address our main research question: Does our latent-variable probe presented in \S\ref{sec:method} outperform previously proposed intrinsic probing methods (\S\ref{sec:results-probe})? Second, we analyze the structure of the most informative m-\bert neurons for the different morphosyntactic attributes we probe for (\S\ref{sec:results-distr}). Finally, we investigate whether knowledge about morphosyntax encoded in neural representations is shared across languages (\S\ref{sec:results-overlap}). In \cref{sec:results-deeper}, we show that our latent-variable probe is flexible enough to support deep neural probes.

\subsection{How Do Our Methods Perform?}
\label{sec:results-probe}

To investigate how the performance of our models compares to existing intrinsic probing approaches, we compare the performance of the \poisson and \condpoisson probes to \lowerbound \citep{dalviWhatOneGrain2019} and \qda \citep{intrinsic}.
We refer to \cref{sec:how-to-compare} for a discussion of the limitations of our method.
\lucas{Eventually (but not now), we should add a footnote that reads ``In concurrent work, Antverg \& Belinkov delve into the limitations of our approach.''}


\begin{table}[t]
\centering
\begin{tabular}{lrrrrr}
\toprule
 & \multicolumn{5}{c}{Number of dimensions} \\
 & $10$ & $50$ & $100$ & $250$ & $500$ \\ \midrule
& \multicolumn{5}{c}{\qda} \\ \midrule
\textsc{C. Poisson}\xspace & {0.50} & {0.58} & {0.70} & {0.99} & {1.00} \\
\poisson & 0.21 & 0.49 & 0.66 & 0.98 & {1.00} \\ \midrule
& \multicolumn{5}{c}{\lowerbound} \\ \midrule
\textsc{C. Poisson}\xspace & {0.99} & {1.00} & {1.00} & {1.00} & {0.98} \\
\poisson & 0.95 & 0.99 & {1.00} & {1.00} & 0.97\\
\bottomrule
\end{tabular}
\caption{Proportion of experiments where \condpoisson (\textsc{C. Poisson}) and \poisson beat the benchmark models \lowerbound and \qda in terms of \NMI. For each of the subset sizes, we sampled 100 different subsets of \bert dimensions at random.}
\label{tab:model_perf}
\end{table}

\begin{table*}[t]
\centering
\resizebox{\textwidth}{!}{
\begin{tabular}{lrrrrrrr}
\toprule
Probe & $10$ & $50$ & $100$ & $250$ & $500$ & $768$ \\ \midrule
\textsc{Cond. Poisson}\xspace & $\mathbf{0.04 \pm 0.03}$ & $\mathbf{0.18 \pm 0.10}$ & $\mathbf{0.31 \pm 0.14}$ & $\mathbf{0.54 \pm 0.17}$ & $\mathbf{0.69 \pm 0.15}$ & $0.71 \pm 0.15$ \\
\poisson & $-0.18 \pm 0.28$ & $0.03 \pm 0.24$ & $0.22 \pm 0.21$ & $0.53 \pm 0.17$ & $\mathbf{0.69 \pm 0.16}$ & $0.71\pm0.19$ \\
\lowerbound & $-0.28 \pm 0.35$ & $-0.18 \pm 0.36$ & $-0.06 \pm 0.35$ & $0.24 \pm 0.33$ & $0.59 \pm 0.21$ & $\mathbf{0.78\pm0.14}$\\
\qda & $-0.15 \pm 0.43$ & $-1.20 \pm 2.82$ & $-3.97 \pm 8.62$ & $-61.70 \pm 186.15$ & $-413.80 \pm 1175.31$ & $-1067.08\pm2420.08$ \\ \midrule
\textsc{Cond. Poisson}\xspace & $0.04\pm0.03$& $0.21\pm0.11$ & $0.35\pm0.16$ & $0.58\pm0.2$ & $0.77\pm0.19$ & $0.74\pm0.16$\\
\poisson & $-0.10\pm0.10$& $0.11\pm0.13$ & $0.28\pm0.17$ & $0.57\pm0.20$ & $0.73\pm0.20$ & $0.76\pm0.18$ \\
\upperbound & $\mathbf{0.10 \pm 0.06}$ & $\mathbf{0.36 \pm 0.16}$ & $\mathbf{0.52 \pm 0.19}$ & $\mathbf{0.70 \pm 0.20}$ & $\mathbf{0.79 \pm 0.17}$ & $\mathbf{0.81\pm0.13}$ \\ 
\bottomrule
\end{tabular}
}
\caption{Mean and standard deviation of \NMI for the \poisson, \condpoisson, \lowerbound~\citep{dalviWhatOneGrain2019} and \qda~\citep{intrinsic} probes for all language--attribute pairs (top) and mean \NMI and standard deviation for the \condpoisson, \poisson and \upperbound for 6 selected language--attribute pairs (bottom). For each subset size considered, we take our averages over 100 randomly sampled subsets of \bert dimensions.}
\label{tab:model_mean_mi}
\end{table*}

\begin{figure}[t]%
    \begin{center}
    \centerline{\includegraphics[width=\columnwidth]{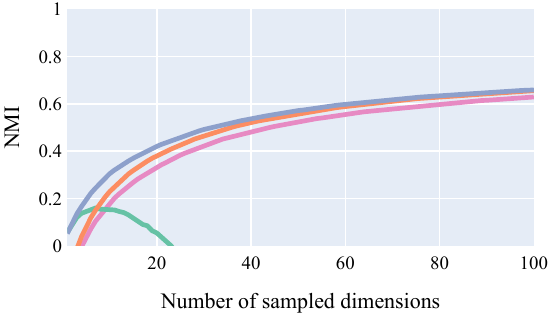}}
    \caption{Comparison of \NMI for the \poissonC, \condpoissonC, \lowerboundC~\citep{dalviWhatOneGrain2019} and \qdaC~\citep{intrinsic} probes. We use the greedy selection approach in \Cref{eq:optimization} to select the most informative dimensions, and average across all language--attribute pairs we probe for.}%
    \label{fig:qda-comparison}%
    \end{center}
\end{figure}


In general, \condpoisson tends to outperform \poisson at lower dimensions, however, \poisson tends to catch up as more dimensions are added. Our results suggest that both variants of our latent-variable model from \cref{sec:method} are effective and generally outperform the \lowerbound baseline as shown in \cref{tab:model_perf}.
The \qda baseline tends to perform similarly to \condpoisson when we consider subsets of 10 dimensions, and it outperforms \poisson substantially.
However, for subsets of size $50$ or more, both \condpoisson and \poisson are preferable.
We believe that the robust performance of \qda in the low-dimensional regimen can be attributed to its ability to model non-linear decision boundaries~\citep[Chapter 4]{murphyMachineLearningProbabilistic2012}.

The trends above are corroborated by a comparison of the mean \NMI (\cref{tab:model_mean_mi}, top) achieved by each of these probes for different subset sizes. However, in terms of accuracy (see \cref{tab:model_mean} in \cref{sec:app-accuracy}), while both \condpoisson and \poisson generally outperform \lowerbound, \qda tends to achieve higher accuracy than our methods.
Notwithstanding, \qda's performance (in terms of \NMI) is not stable and can yield low or even negative mutual information estimates across all subsets of dimensions. Adding a new dimension can never decrease the mutual information, so the observable decreases occur because the generative model deteriorates upon adding another dimension, which validates \citeauthor{intrinsic}'s claim that some dimensions are not adequately modeled by the Gaussian assumption.
While these results suggest that \qda may be preferable if performing a comparison based on accuracy, the instability of \qda when considering \NMI suggests that this edge in terms of accuracy comes at a hefty cost in terms of calibration~\citep{guo2017calibration}.\footnote{While accuracy only cares about whether predictions are correct, \NMI penalizes miscalibrated predictions since it is proportional to the negative log likelihood~\citep{guo2017calibration}.}
\lucas{We have to be careful here; our accuracy table doesn't show that clear of an improvement. We should probably elaborate on this.}\karolina{Tried to rewrite it a bit, what do you think?}\lucas{I think we need to address this head-on, especially since NMI is a ``strange'' metric compared to accuracy. What do you think?}

Further, we compare the \poisson and \condpoisson probes to the \upperbound baseline. This is expected to be the highest performing since it is re-trained for \emph{every} subset under consideration and indeed, this assumption is confirmed by the results in \cref{tab:model_mean_mi} (bottom). 
The difference between our probes' performance and the \upperbound baseline's performance can be seen as the cost of sharing parameters across all subsets of dimensions, and an effective intrinsic probe should minimize this.

We also conduct a direct comparison of \lowerbound, \qda, \poisson, and \condpoisson when used to identify the most informative subsets of dimensions.
The average MI reported by each model across all \XX morphosyntactic language--attribute pairs is presented in \cref{fig:qda-comparison} (see \cref{fig:qda-comparison-acc} in the Appendix for the accuracy comparison).
On average, \condpoisson offers comparable performance to \qda at low dimensionalities for both \NMI and accuracy, though the latter tends to yield a slightly higher (and thus a tighter) bound on the MI.
However, as more dimensions are taken into consideration, our models vastly outperform \qda.
Our models perform comparably at high dimensions, but \condpoisson performs slightly better for 1--20 dimensions.
\poisson outperforms \lowerbound at high dimensions, and \condpoisson outperforms \lowerbound for all dimensions considered. 
These effects are less pronounced for accuracy, which we believe to be due to accuracy's insensitivity to a probe's confidence in its prediction. 
Finally, while \condpoisson achieves a tighter bound on \NMI than \poisson, we recommend the \poisson probe for larger experimental setups due to its computational efficiency.   


\subsection{Information Distribution}
\label{sec:results-distr}
\begin{figure}[t]
\begin{center}
    \includegraphics[width=\columnwidth]{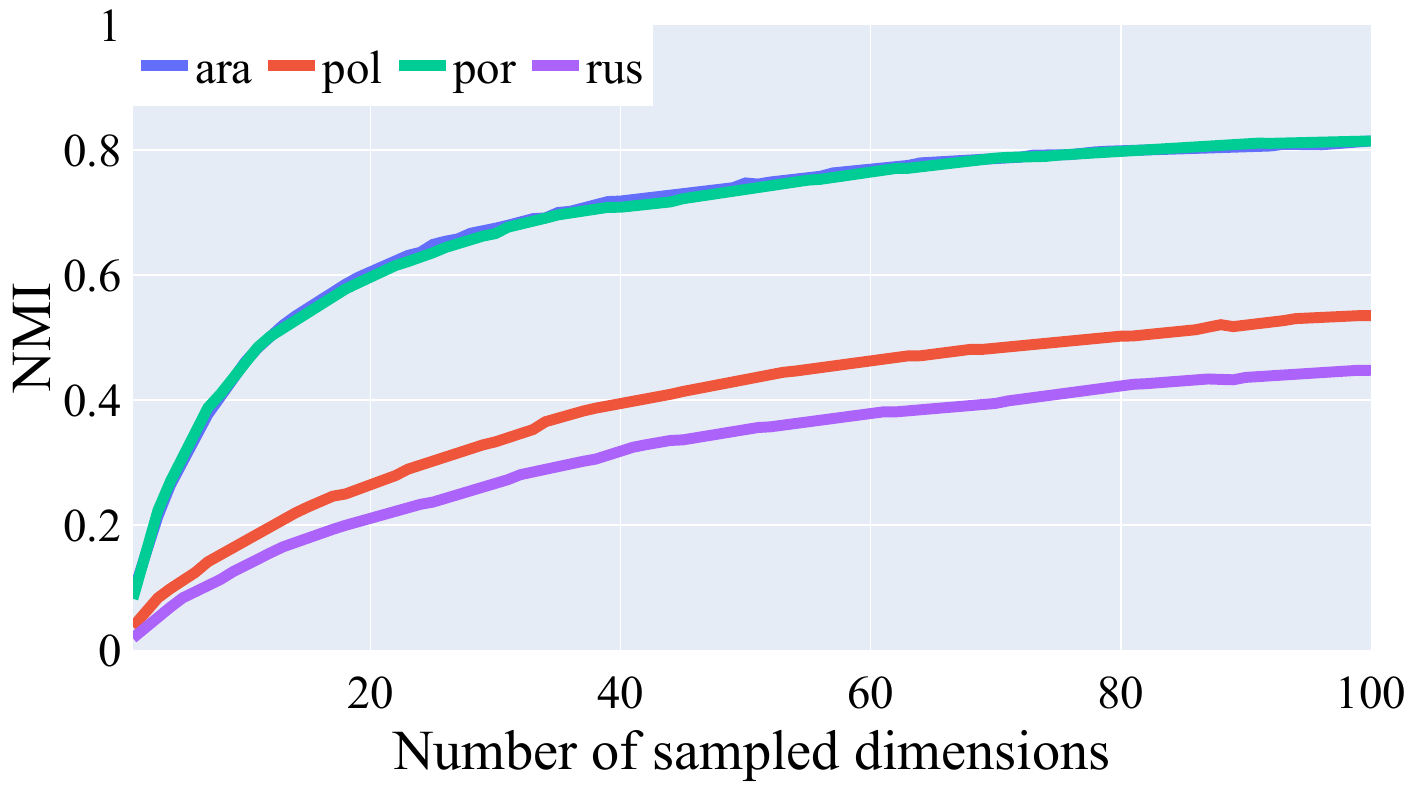}
    \caption{Comparison of the average \NMI for gender dimensions in \bert for each of the available languages. We use the greedy selection approach in \Cref{eq:optimization} to select the most informative dimensions, and average across all language–attribute pairs we probe for.}
    \label{fig:gen-comparison}
\end{center}
\end{figure}

We compare performance of the \condpoisson probe for each attribute for all available languages in order to better understand the relatively high \NMI variance across results (see \cref{tab:model_mean_mi}).
In \cref{fig:gen-comparison}, we plot the average \NMI for gender\lucas{Hmmm.. should we do this for the greedy method instead? The current interpretation is that any subset of dimensions will encode less information about gender if there are more classes.}\karolina{Yes, already changed to greedy selection. Need to check what is the reason for the peak around 3-4 dimensions.} and observe that languages with two genders present (Arabic and Portuguese) achieve higher performance than languages with three genders (Russian and Polish) which is an intuitive result due to increased task complexity. Further, we see that the slopes for both Russian and Polish are flatter, especially at lower dimensions. This implies that the information for Russian and Polish is more dispersed and more dimensions are needed to capture the typological information.    

\subsection{Cross-Lingual Overlap}
\label{sec:results-overlap}
\begin{figure}[t]
    \centering
    \includegraphics[width=\columnwidth]{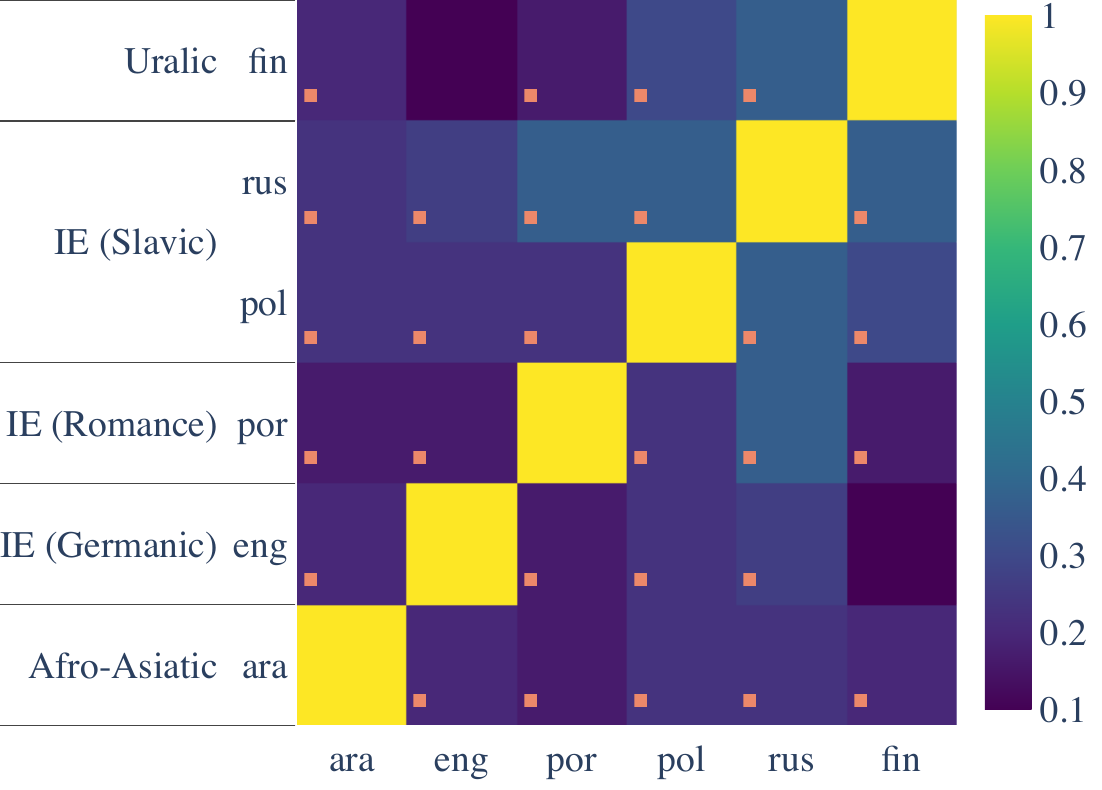}
    \caption{The percentage overlap between the top 30 most informative number dimensions in \bert for the probed languages. Statistically significant overlap, after Holm--Bonferroni family-wise error correction~\citep{holmSimpleSequentiallyRejective1979}, with $\alpha = 0.05$, is marked with an orange square.
    }
    \label{fig:heatmap_gender}
\end{figure}

We compare the most informative m-\bert dimensions recovered by our probe across languages and find that, in many cases, the same set of neurons express the same morphosyntactic phenomena across languages.
For example, we find that Russian, Polish, Portuguese, English, and Arabic have statistically significant overlap in the top 30 most informative neurons for number~(\cref{fig:heatmap_gender}). Similarly, we observe presence of statistically significant overlap for gender~(\cref{fig:heatmap-gender-case}, left). 
This effect is particularly strong between Russian and Polish, where we find statistically significant overlap between top-30 neurons for case~(\cref{fig:heatmap-gender-case}, right).
These results indicate that \bert may be leveraging data from other languages to develop a cross-lingually entangled notion of morpho-syntax~\citep{intrinsic} and that this effect may be particularly strong between typologically similar languages.\footnote{Recently, both \citet{stanczak-etal-2022-neurons}, who utilize the \poisson probe, and \citet{antverg2021pitfalls} find evidence supporting a similar phenomenon.}

%

\section{Related Work}
\label{sec:related-work}

A growing interest in interpretability has led to a flurry of work in assessing what pre-trained representations know about language.
To this end, diverse methods have been employed, such as the construction of challenge sets that evaluate how well representations model particular phenomena~\citep{linzenAssessingAbilityLSTMs2016,gulordavaColorlessGreenRecurrent2018,goldbergAssessingBERTSyntactic2019,goodwinProbingLinguisticSystematicity2020},
and visualization methods~\citep{kadarFixed,conf/uai/RethmeierSA20}.
Work on probing comprises a major share of this endeavor~\citep{belinkovAnalysisMethodsNeural2019,belinkov2021probing}.
This has taken the form of focused studies on particular linguistic phenomena~\citep[e.g., subject-verb number agreement,][]{giulianelliHoodUsingDiagnostic2018}
to broad assessments of contextual representations in a wide array of tasks~\citep[\emph{inter alia}]{sahinLINSPECTORMultilingualProbing2020,tenneyWhatYouLearn2018,conneauWhatYouCan2018, ravichander-etal-2021-probing,geva2022transformer}.

Efforts have ranged widely, but most of these focus on extrinsic rather than intrinsic probing.
Most work on the latter has focused primarily on ascribing roles to individual neurons through methods such as visualization~\citep{karpathyVisualizingUnderstandingRecurrent2015,liVisualizingUnderstandingNeural2016} and ablation~\citep{liUnderstandingNeuralNetworks2017}.
For example, recently \citet{lakretzEmergenceNumberSyntax2019} conduct an in-depth study of how LSTMs \citep{hochreiterLongShortTermMemory1997} capture subject--verb number agreement, and identify two units largely responsible for this phenomenon.

More recently, there has been a growing interest in extending intrinsic probing to collections of neurons.
\citet{bauIdentifyingControllingImportant2019} utilize unsupervised methods to identify important neurons and then attempt to control a neural network's outputs by selectively modifying them.
\citet{bauUnderstandingRoleIndividual2020} pursue a similar goal in a computer vision setting but ascribe meaning to neurons based on how their activations correlate with particular classifications in images and are able to control these manually with interpretable results.
Aiming to answer questions on interpretability in computer vision and natural language inference, \citet{mu2021compositional} develop a method to create compositional explanations of individual neurons and investigate abstractions encoded in them. \citet{vigInvestigatingGB2020} analyze how information related to gender and societal biases is encoded in individual neurons and how it is being propagated through different model components.



\section{Conclusion}
\label{sec:conclusion}

In this paper, we introduce a new method for training intrinsic probes. 
We construct a probing classifier with a subset-valued latent variable and demonstrate how the latent subsets can be marginalized using variational inference. We propose two variational families, based on common sampling designs, to model the posterior over subsets: Poisson and conditional Poisson sampling.
We demonstrate that both variants outperform our baselines in terms of mutual information and that using a conditional Poisson variational family generally gives optimal performance.
Next, we investigate information distribution for each attribute for all available languages.
Finally, we find empirical evidence for overlap in the specific neurons used to encode morphosyntactic properties across languages.


\section{Acknowledgements}
This research was co-funded by a DFF Research Project 1 under grant agreement No 9130-00092B.

\bibliography{tacl2018}

\clearpage

\appendix


\section{List of Probed Morphosyntactic Attributes}
\label{sec:app-attributes}
The \XX language--attribute pairs we probe for in this work are listed below:
\begin{itemize}
    \item \textbf{Arabic}: Aspect, Case, Definiteness, Gender, Mood, Number, Voice
    \item \textbf{English}: Number, Tense
    \item \textbf{Finnish}: Case, Number, Person, Tense, Voice
    \item \textbf{Polish}: Animacy, Case, Gender, Number, Tense
    \item \textbf{Portuguese}: Gender, Number, Tense
    \item \textbf{Russian}: Animacy, Aspect, Case, Gender, Number, Tense, Voice
\end{itemize}


\section{How Do Deeper Probes Perform?}
\label{sec:results-deeper}

Multiple papers have promoted the use of linear probes~\citep{tenneyWhatYouLearn2018,liuLinguisticKnowledgeTransferability2019}, in part because they are ostensibly less likely to memorize patterns in the data~\citep{zhang-bowman-2018-language,hewittDesigningInterpretingProbes2019}, though this is subject to debate~\citep{voita-titov-2020-information,pimentelInformationtheoreticProbingLinguistic2020}.
Here we verify our claim from \cref{sec:method} that our probe can be applied to any kind of discriminative probe architecture as our objective function can be optimized using gradient descent.

We follow the setup of \citet{hewittDesigningInterpretingProbes2019}, and test MLP-1 and MLP-2 \condpoisson probes alongside a linear \condpoisson probe. 
The MLP-1 and MLP-2 probes are multilayer perceptrons (MLP) with one and two hidden layer(s), respectively, and Rectified Linear Unit~\citep[ReLU;][]{nairRectifiedLinearUnits2010} activation function.

In \cref{tab:nonlinear-mi}, we can see that our method not only works well for deeper probes but also outperforms the linear probe in terms of \NMI. 
We note that the difference in performance between MLP-1 and MLP-2 is negligible.

\section{Supplementary Results}
\label{sec:app-accuracy}

\begin{figure}[t]%
    \begin{center}
    \includegraphics[width=\columnwidth]{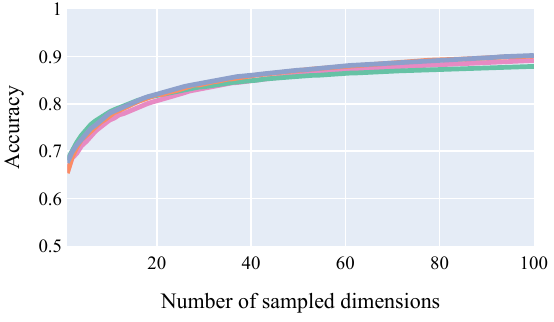}
    \caption{Comparison of the \poissonC, \condpoissonC, \lowerboundC~\citep{dalviWhatOneGrain2019} and \qdaC~\citep{intrinsic} probes. We use the greedy selection approach in \Cref{eq:optimization} to select the most informative dimensions, and average across all language--attribute pairs we probe for.}%
    \label{fig:qda-comparison-acc}%
    \end{center}
\end{figure}

\Cref{tab:model_mean} compares the accuracy of our two models, \poisson and \condpoisson, to the \lowerbound, \qda and \upperbound baselines.
The table reflects the trend observed in \cref{tab:model_mean_mi}: \poisson and \condpoisson generally outperform the \lowerbound baseline. However, \qda achieves higher accuracy with exception of a high-dimension regimen. In \cref{fig:qda-comparison-acc}, the accuracy reported by each model across all \XX morphosyntactic language--attribute pairs is presented.

\Lucas{This table suggests that \qda is better than our probe! We should probably explain this in the text.}


\begin{table*}[ht]
\centering
\begin{tabular}{lrrrrr}
\toprule
Probe & $10$ & $50$ & $100$ & $250$ & $500$ \\ \midrule
\textsc{Linear Cond. Poisson}\xspace & $0.04\pm0.03$ & $0.21\pm0.11$ & $0.35\pm0.15$ & $0.59\pm0.19$ & $0.74\pm0.18$\\
\textsc{MLP-1} & $\mathbf{0.06\pm0.05}$ & $0.26\pm0.13$ & $0.43\pm0.16$ & $0.67\pm0.17$ & $\mathbf{0.80\pm0.14}$ \\
\textsc{MLP-2} & $\mathbf{0.06\pm0.05}$ & $\mathbf{0.27\pm0.13}$ & $\mathbf{0.44\pm0.17}$ & $\mathbf{0.68\pm0.17}$ & $\mathbf{0.80\pm0.14}$\\
\bottomrule
\end{tabular}
\caption{Mean and standard deviation of the NMI for the \textsc{linear} \condpoisson probe to non-linear \textsc{MLP-1} and \textsc{MLP-2} \condpoisson probes for selected language-attribute pairs. For each of the subset sizes, we sampled 100 different subsets of \bert dimensions at random.}
\label{tab:nonlinear-mi}
\end{table*}

\begin{table*}[ht]
\centering
\resizebox{\textwidth}{!}{
\begin{tabular}{lrrrrrr}
\toprule
Probe & $10$ & $50$ & $100$ & $250$ & $500$ & $768$ \\ \midrule
\textsc{Cond. Poisson}\xspace & $0.66\pm0.15$ & $0.73\pm0.13$ & $0.78\pm0.11$ & $0.86\pm0.08$ & $\mathbf{0.92\pm0.06}$ & $0.93\pm0.05$\\
\poisson & $0.62\pm0.15$ & $0.70\pm0.13$ & $0.77\pm0.12$ & $0.86\pm0.08$ & $\mathbf{0.92\pm0.06}$ & $0.94\pm0.04$ \\
\lowerbound & $0.51\pm0.15$ & $0.59\pm0.15$ & $0.65\pm0.14$ & $0.77\pm0.12$ & $0.88\pm0.08$ & $\mathbf{0.95\pm0.04}$\\
\qda & $\mathbf{0.69\pm0.14}$ & $\mathbf{0.80\pm0.11}$ & $\mathbf{0.84\pm0.09}$ & $\mathbf{0.88\pm0.08}$ & $0.88\pm0.08$ & $0.87\pm0.1$\\ \midrule
\textsc{Cond. Poisson}\xspace & $0.55\pm0.1$ & $0.65\pm0.13$ & $0.72\pm0.12$ & $0.83\pm0.10$ & $0.90\pm0.08$ & $0.93\pm0.06$\\
\poisson & $0.51\pm0.13$ & $0.63\pm0.14$ & $0.72\pm0.12$ & $0.83\pm0.10$ & $0.90\pm0.08$ & $0.93\pm0.07$ \\
\upperbound & $\mathbf{0.58\pm0.12}$ & $\mathbf{0.75\pm0.12}$ & $\mathbf{0.80\pm0.10}$ & $\mathbf{0.89\pm0.08}$ & $\mathbf{0.93\pm0.06}$ & $\mathbf{0.94\pm0.05}$\\ 
\bottomrule
\end{tabular}
}
\caption{Mean and standard deviation of accuracy for the \poisson, \condpoisson, \lowerbound~\citep{dalviWhatOneGrain2019} and \qda~\citep{intrinsic} probes for all language--attribute pairs (above) and for the \condpoisson, \poisson and \upperbound for 6 selected language--attribute pairs (below) for each of the subset sizes. We sampled 100 different subsets of \bert dimensions at random.}
\label{tab:model_mean}
\end{table*}

\begin{figure*}[t]
\begin{center}
\includegraphics[width=\columnwidth]{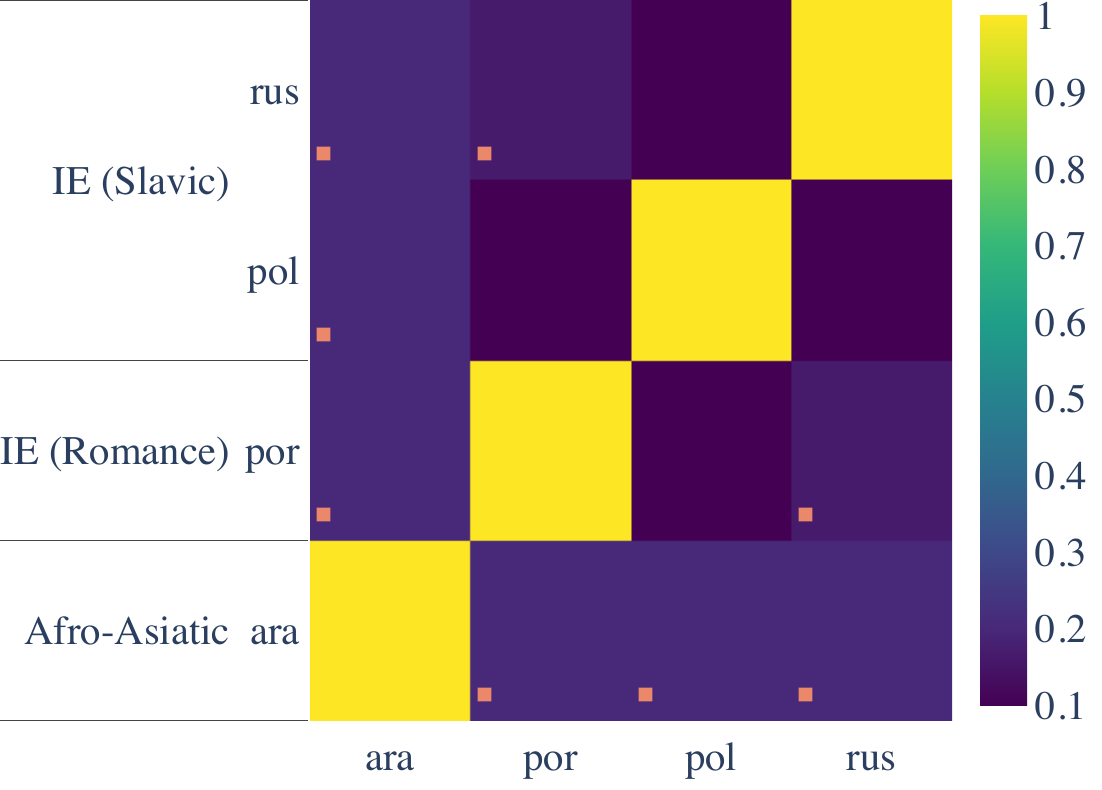}
\hfill
\includegraphics[width=\columnwidth]{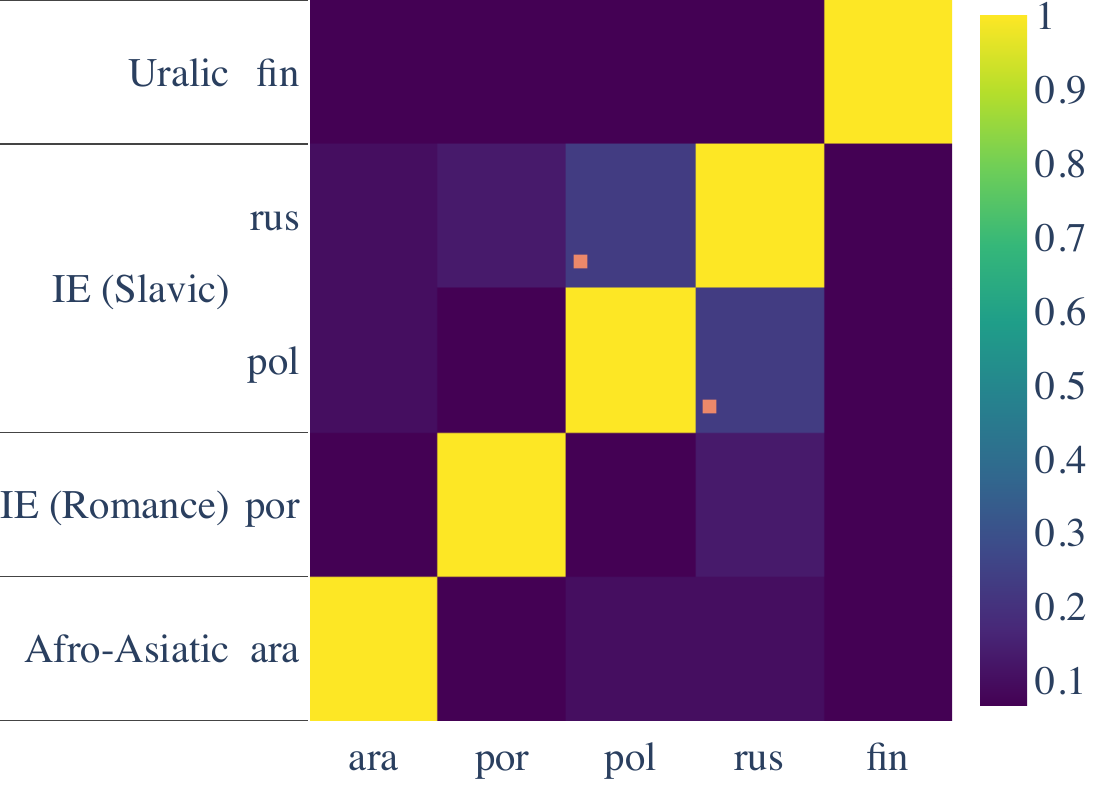}
\caption{The percentage overlap between the top-30 most informative gender (left) and case (right) dimensions in \bert for the probed languages. Statistically significant overlap, after Holm--Bonferroni family-wise error correction~\citep{holmSimpleSequentiallyRejective1979}, with $\alpha = 0.05$, is marked with an orange square.}
\label{fig:heatmap-gender-case}
\end{center}
\end{figure*}

\end{document}